# LncRNA-disease association prediction method based on heterogeneous information completion and convolutional neural network

Wen-Yu Xi. Juan Wang, Yu-Lin Zhang, Jin-Xing Liu,Yin-Lian Gao

*Abstract*—The emerging research shows that lncRNA has crucial research value in a series of complex human diseases. Therefore, the accurate identification of lncRNA-disease associations (LDAs) is very important for the warning and treatment of diseases. However, most of the existing methods have limitations in identifying nonlinear LDAs, and it remains a huge challenge to predict new LDAs. In this paper, a deep learning model based on a heterogeneous network and convolutional neural network (CNN) is proposed for lncRNA-disease association prediction, named HCNNLDA. The heterogeneous network containing the lncRNA, disease, and miRNA nodes, is constructed firstly. The embedding matrix of a lncRNA-disease node pair is constructed according to various biological premises about lncRNAs, diseases, and miRNAs. Then, the low-dimensional feature representation is fully learned by the convolutional neural network. In the end, the XGBoot classifier model is trained to predict the potential LDAs. HCNNLDA obtains a high AUC value of 0.9752 and AUPR of 0.9740 under the 5-fold cross-validation. The experimental results show that the proposed model has better performance than that of several latest prediction models. Meanwhile, the effectiveness of HCNNLDA in identifying novel LDAs is further demonstrated by case studies of three diseases. To sum up, HCNNLDA is a feasible calculation model to predict LDAs.

## I. Introduction

WITH the implementation of the human genome project, researchers have found that only a small part of the genes can encode a protein, and most of the genome sequence is non-protein-coding region [1-3]. Among them, lncRNAs are closely related to multiple biological processes, which have significant functions in chromatin modification, transcriptional silencing and activation, chromosome modification [4, 5]. By using machine learning in biological research, researchers have discovered many potential lncRNA-disease associations (LDAs) through various experiments and various data analysis methods [6]. It is worth noting that up to now, there are some gaps in the research on the function of lncRNA due to many restrictions [7-10]. Therefore, the research focusing on identifying LDAs can help humans prevent and treat certain diseases to a great extent.

At present, many biological experiments have verified that the abnormality of lncRNA is potentially related to complex diseases, including nervous system diseases [11], leukemia [12], cardiovascular diseases [13] and Alzheimer's disease[14]. Therefore, the research on lncRNAs has become one of the popular fields. Just like PVT1 is a kind of lncRNA, which is closely related to cancers such as gastric cancer[15], pancreatic cancer [16] and prostate cancer [17]. In recent years, several public databases for storing LDAs have been available for biologists to study their function and phenotypic characteristics [18-21]. However, the determination of LDAs through biological experiments requires large-scale manpower and material resources. So far, only few lncRNAs have good functional characteristics. In summary, it is essential to develop a viable predictive model to verify the potential LDAs.

With the progress of biotechnology, many computational models for predicting LDAs have been developed [22, 23]. These models have been successfully verified by biological experiments [24], and they can be roughly classified into three categories. The first type of method is to use machine learning method to identify potential LDAs, which is essentially based on similarity measure and matrix factorization. Chen *et al.*

This work was partly supported by the National Natural Science Foundation of China (Grant Nos. 62172254).

Wen-Yu Xi is with the School of Information Science and Engineering, Qufu Normal University, Rizhao, 276826, China (e-mail: xiwenyu1007@126.com).

proposed the method, called LRLSLDA, to predict unknown LDAs by means of Laplacian regularized least squares [25], and improved the accuracy of predicting the LDAs. Chen *et al.* introduced a method to find unknown LDAs by integrating all kinds of similarity information using KATZ measure (KATZLDA) [26]. This approach can predict LDAs without known association information. Ping *et al.* [27] established a bipartite graph based on the known LDAs, and analyzed its properties, so as to predict the potentially lncRNA-disease pairs. Liu *et al.* identified the potential LDAs by adding graph regularization and weight matrix to the collaborative matrix factorization (WGRCMF)[28], which greatly improved the accuracy of the prediction. Zhao *et al.* integrated multiple isomorphic and heterogeneous networks, constructed the multi-layer network, and carried out the restarted random walk on the network to predict LDAs (MHRWR) [29]. However, these methods fail to combine with a variety of multi-source information related to lncRNAs, such as genes, proteins.

The approaches of the second category incorporate multiple data sources related to lncRNAs and diseases. Chen *et al.* fused multiple similarities of genes, lncRNAs and diseases to establish a novel model called ILDMSF [30], which improved the prediction results compared with the methods of single similarity. Lan *et al.* integrated various similarities between lncRNAs and diseases, and used SVM model to identify candidate LDAs [31]. Zhang *et al.* used flow propagation algorithm to infer potential LDAs by constructing several lncRNA-protein-disease association networks [7]. Similarly, Fan *et al.* introduced protein information related to lncRNAs and diseases, and used the RWR algorithm to extract the topological properties between the association networks to identify potential LDAs [32]. Although the above methods make full use of multi-source information, it is difficult to explore the deep relationship between the converged networks.

The approaches of the third category are to use deep learning algorithm to identify potential LDAs. With the emergence of various deep models, biologists began to use them to explore the potential LDAs, mine their deep characteristics and improve the prediction performance. Guo *et al.* [33] learned the deep characteristics between lncRNAs and diseases by using a deep autoencoder, and used rotation forest for prediction. Xuan *et al.* [34] established a dual convolutional neural network model to measure disease-related lncRNAs from a variety of biological information. Although existing models are effective in predicting potential LDAs, there is still a possibility to improve the accuracy.

In our research, a novel framework for predicting LDAs is proposed, called HCNNLDA. Our method has the following three advantages:

- Integration of heterogeneous multi-source data, including miRNA-disease associations (MDAs), LDAs and miRNA-lncRNA associations (MLAs), to generate high-dimensional feature representations of lncRNA and disease.
- Convolutional neural network (CNN) is used to learn the original eigenvector in low dimension to obtain the optimal subspace.
- Considering the possible nonlinear relationship of the features obtained after dimensionality reduction, XGBoot is used to predict the potential LDAs.

The overall framework of HCNNLDA is shown in Figure 1. The HCNNLDA model was validated using 5-fold cross-validation (CV), and the experimental results show that HCNNLDA has better performance than the previous prediction models.

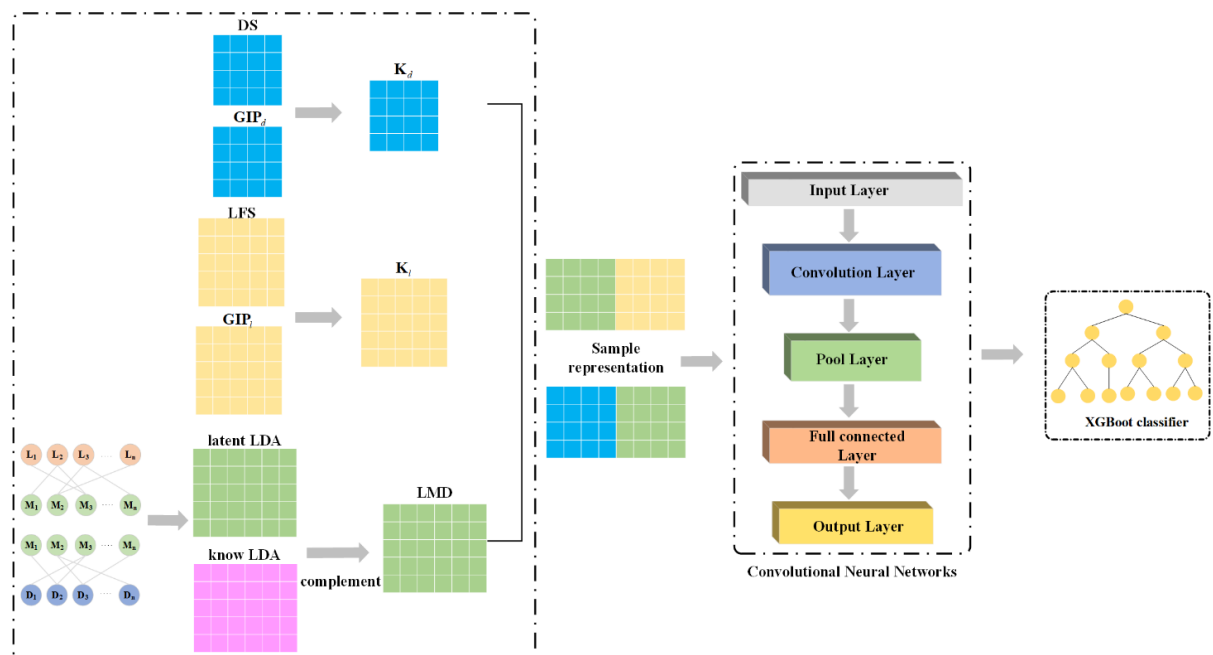

Figure 1. The overall framework of HCNNLDA.

## II. MATERIALS

### A. Datasets

The datasets used in this paper encompass experimentally supported Long non-coding RNA-Disease Associations (LDAs), MiRNA-Disease Associations (MDAs), and MiRNA-LncRNA Associations (MLAs). The LDAs supported by 2697 experiments is initially downloaded from the Lnc2Cancer [35], LncRNADisease database [36]. A total of 13,562 experimentally supported MDAs were sourced from the HMDD (v2.0) database [37]. Additionally, 1,002 MLAs were downloaded from the StarBase database [38]. The processed data included 240 lncRNAs, 495 miRNAs and 412 diseases. Here, adjacency matrix $Y_{ld}$ is used to represent LDAs, if lncRNA $l_i$ is related to disease $d_j$, $Y_{ld}(i,j) = 1$; otherwise, $Y_{ld}(i,j) = 0$. Similarly, MDAs and MLDs are represented by adjacency matrices $Y_{md}$ and $Y_{ml}$, respectively. The dataset details are shown in TABLE I.

TABLE I
The dataset used in the experiment

| Datasets | Matrix size | Associations |
|---|---|---|
| LDAs (LD) | 240×412 | 2697 |
| MDAs (MD) | 495×412 | 13562 |
| MLAs (ML) | 240×495 | 1002 |

### B. Disease Semantic Similarity Matrix

The disease semantic similarity is calculated by disease directed acyclic graph (DAG). For disease $n$, it can be represented $DAG(d) = (N(d), T(d))$, where $N(d)$ is the set of $n$ and all its ancestor nodes. Therefore, the semantic contribution value of node $n$ to disease $d$ is defined as:

$$D1_d(n) = \begin{cases} 1 & \text{if } n = d, \\ \max\{\Delta * D_d(\hat{n}) | \hat{n} \in \text{children of } n\} & \text{if } n \neq d. \end{cases}$$

where $\Delta$ represents a semantic contribution factor. According to previous research and experience, better experimental results can be obtained when $\Delta = 0.5$ [28]. Therefore, the semantic contribution value of disease $d$ is represented as:

$$DV1(d) = \sum_{d \in N(d)} D_d(d),$$

At last, the similarity between the two diseases is obtained from Equation (3):

$$\mathbf{DS}_l(A, B) = \frac{\sum_{t \in N(A) \cap N(B)} (D_A(t) + D_B(t))}{DV(A) + DV(B)}$$

By calculating these semantic contributions for all nodes in the DAG, we can construct a semantic similarity matrix for diseases. This matrix allows for the comparison of diseases based on their underlying semantic structures, facilitating better understanding and analysis of disease relationships and characteristics.

### C. LncRNA Functional Similarity

In a previous study by Chen *et al.*, the functional similarity of lncRNAs was calculated to understand their roles in various diseases. The similarity between each pair of lncRNAs is determined based on the diseases they are associated with. For each pair of lncRNAs, their similarity is defined as:

$$LFS(i,j) = \frac{\sum_{d \in D(l_j)} S(d, D(l_i)) + \sum_{d \in D(l_i)} S(d, D(l_j))}{m + n},$$

$$S(d, D(l_i)) = \max_{d_1 \in D(l_i)} (S(d, d_1))$$

where m and n represent the number of diseases related to lncRNA $l_i$ and $l_j$, and $D(l)$ represents the disease related to lncRNA $l$.

By employing this method, researchers can effectively compare lncRNAs and gain insights into their potential functional roles in disease mechanisms, contributing to a deeper understanding of lncRNA biology and its implications in medical research.

### D. Gaussian interaction profile kernel similarity of disease and lncRNA

GIP kernel similarity is a commonly used similarity measurement method in the LDA prediction process. Based on the assumption that functionally similar lncRNAs are more likely to be associated with similar diseases, GIP kernel similarity describes the similarity information between lncRNAs and diseases. The GIP kernel similarity for lncRNAs and diseases is mathematically defined as follows:

$$\mathbf{GIP}_l(l_p, l_q) = \exp(-\gamma_m \|\mathbf{Y}(l_p) - \mathbf{Y}(l_q)\|^2),$$

$$\mathbf{GIP}_d(d_p, d_q) = \exp(-\gamma_d \|\mathbf{Y}(d_p) - \mathbf{Y}(d_q)\|^2).$$

Here, $l_p$ and $l_q$ represent two different lncRNAs, $d_p$ and $d_q$ represent two different diseases, $\mathbf{Y}(l_p)$ denotes the p-th row vector in the adjacency matrix Y, $\mathbf{Y}(d_q)$ denotes the q-th column vector in the adjacency matrix Y. Additionally, $\gamma_l$ and $\gamma_d$ represent the parameters controlling the kernel bandwidth, which can be expressed as:

$$\gamma_l = \gamma_l' / \left(\frac{1}{n_l} \sum_{i=1}^{n_l} \|\mathbf{Y}(l_p)\|^2\right),$$

$$\gamma_d = \gamma_d' / \left(\frac{1}{n_d} \sum_{i=1}^{n_d} \|\mathbf{Y}(d_q)\|^2\right).$$

## III. METHODOLOGY

### A. Fusing different similarities for lncRNA and disease

In this paper, we used the maximum value method to merge lncRNA Gaussian interaction profile kernel similarity and lncRNA functional similarity into LFS similarity and fuse disease Gaussian interaction profile kernel similarity and disease Semantic similarity into DS similarity:

$$LFS = \begin{cases} GIP_l(l_i, l_j) & if\, GIP_{lnc}(l_i, l_j) \geq LFS(i,j) \\ LFS(i,j) & otherwise \end{cases}$$

$$DS = \begin{cases} GIP_d(d_i, d_j) & if\, GIP_d(d_i, d_j) \geq DS(i,j) \\ DS(i,j) & otherwise \end{cases}$$

## B. Geometric complement for lncRNA- disease associations matrix

Inspired by previous methods , we derived a potential LDA matrix from the previous data source. Specifically, we multiplied the LMI matrix by the MDA matrix and then divided the [i, j]-th element of the resulting product by the i-th row sum of the LMI matrix and the j-th column sum of the MDA matrix, as represented by Equation :

$$\text{LMD}(i, j) = \frac{LM(i,:) \cdot MD(:,j)}{\|LM(i,:)\|_1 + \|MD(:,j)\|_1}.$$

The fusion matrix of LDA was obtained by taking the maximum value between the potential LDA matrix computed above and the original LDA matrix at each [i,j]-th position.

$$LD_{new}(i,j) = max(LD(i,j), \text{LMD}(i,j))$$

For the geometric complementary matrix obtained in the previous section, where each row of the matrix represents the feature vector of an lncRNA and each column represents the feature vector of a disease, a feature fusion strategy was adopted to further enhance the model's performance. Specifically, the i-th row of the geometric complementary matrix of lncRNAs is combined with the i-th row of the similarity fusion matrix to form a new i-th row feature vector for the lncRNA. Similarly, the j-th column of the geometric complementary matrix of diseases is combined with the j-th column of the disease similarity fusion matrix to form a new j-th feature vector for the disease. This fusion method effectively leverages the information from both different feature matrices, enhancing the model's performance and robustness. Ultimately, after feature fusion, we obtain the feature representations for lncRNAs and diseases.

## C. Feature extraction based on CNN

Recently, convolutional neural networks are not only widely used in image processing [39, 40], but also attracted more and more attention in biomedical field. Given CNN can effectively process all kinds of original image data[41], multiple convolutions check original feature vectors in CNN's convolution layer are used for convolution operation to obtain the feature representation of lncRNA and disease. The i-th row in the matrix is defined as the i-th disease:

$$\lambda_i = \sigma(\lambda_{i-1} \otimes W_i + b_i),$$

where $W_i$ represents the weight matrix of the i-th layer; $\otimes$ represents convolution operation; $b_i$ is the offset vector; $\sigma(x)$ is the activation function. Next, in the pooling layer $\lambda_i$, the pooling process is as follows:

$$\lambda_i = subsampling(\lambda_{i-1}),$$

The convolution neural network is formed by alternating the collection of convolution layer and pooling layer, and then the feature selection is carried out through the pooling layer. The extracted features are learned by fully connected layer and probability distribution F. Here, the original input matrix $\lambda_0$ is mapped to the new characteristic expression F by CNN.

$$F(i) = Map(A = \lambda_i | \lambda_0; (W, b)), \qquad (10)$$

where F is the characteristic expression, $\lambda_i$ represents the i-th label classification, and $\lambda_0$ represents the original input matrix. The minimization of the loss function of the neural network $S(W, b)$ is taken as the training target. At the same time, in order to avoid overfitting, the loss function $Q(W, b)$ is controlled by using a norm, in which the parameter $\delta$ is used to control the intensity of the fitting.

$$Q(W,b) = S(W,b) + \frac{\delta}{2} W^T W,$$

The convolutional neural network is optimized by gradient descent algorithm, and the parameters (W, b) were updated. The intensity of back propagation was controlled by the learning rate $\theta$.

$$W_i = W_i - \theta \frac{\partial Q(W,b)}{\partial W_i},$$

$$b_i = b_i - \theta \frac{\partial Q(W,b)}{\partial b_i},$$

After several experiments to optimize the parameters, the kernel size of the convolution layer is finally selected as 16×16, and the kernel size of the sub-sampling layer is 2×2. Binary_crossentropy is used as a loss function.

## D. Extreme gradient boosting

XGBoost is a gradient boosting decision tree, which integrates many tree models to form a strong classifier. XGBoost has been widely used in the field of biology and achieved good results [42]. The objective function of XGBoost algorithm is defined as:

$$obj(\theta) = \sum_{i}^{n} l(y_i, \hat{y}_i) + \sum_{t=1}^{T} \Omega(f_t),$$

where $l$ is the loss function, and $f_t$ is the t-th tree. The second-order Taylor series of $L$ at the t-th iteration is obtained:

$$L^{(t)} \simeq \sum_{i=1}^{k} \left[ l(y_i, \hat{y}^{(t-1)}) + g_i f_t(x_i) + \frac{1}{2} h_i f_t^2(x_i) \right] + \Omega f_t.$$

where $g_i$, $h_i$ represents the first and second derivative of each sample. The XGBoost package can be downloaded at https://github.com/dmlc/xgboost. In the whole experiment, the default parameter set of XGBoost training package implemented by Chen *et al.* [43]. And the number of Boosting trees is set to 500, and the maximum dedpth is 15. The parameters of the specific results are shown in Figure 2 and Figure 3.

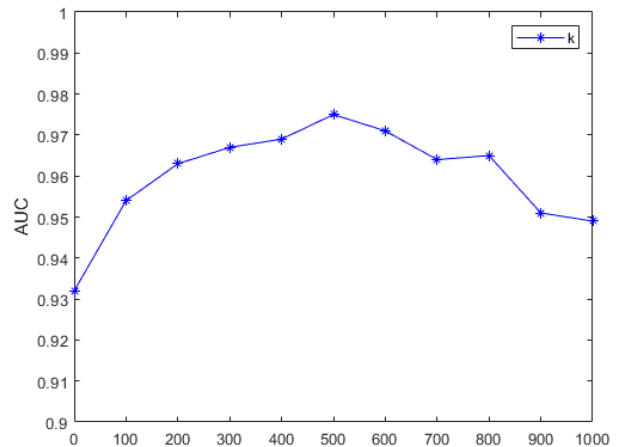

Figure 2. Comparing the output of different decision trees k.

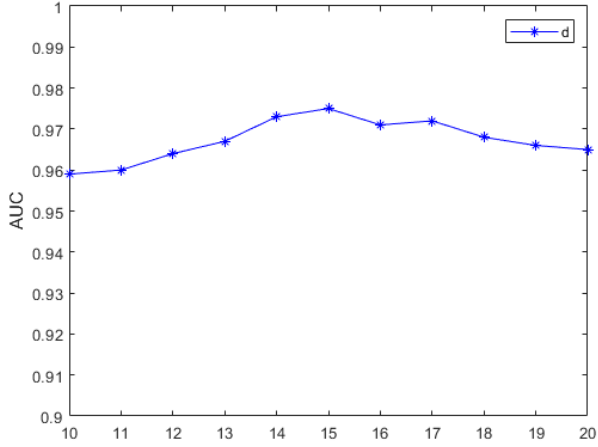

Figure 3. Comparing the output of different depth d.

## IV. RESULTS AND DISCUSSIONS

### A. Evaluation Metrics

In order to prove the ability of HCNNLDA, a series of experiments are carried out. Firstly, four different classifiers such as SVM, random forest (RF), gradient Boosting (GBDT), and AdaBoost are compared with HCNNLDA. Secondly, compare our method with the latest experiment under 5-fold CV.

For the purpose of proving the performance of HCNNLDA, receiver operating characteristic (ROC) curve is used as the measurement index to calculate the area under ROC (AUC) value. Meanwhile, three evaluation indexes, including Accuracy (ACC), F1-Score (F1) and Precision (Pre) are also used, which are calculated as:

$$Acc = \frac{TP+TN}{TP+FN+TN+FP},$$

$$\Pr e = \frac{TP}{TP+FP},$$

$$F1 = \frac{2TP}{2TP+FP+FN},$$

where TP, FP, TN, FN are the number of true positives, false positives, true negatives, and false negatives in the sample matrix, respectively.

### B. Comparison with different classifiers

Different classifiers are compared, and the ROC curve is shown in Figure 2. The comprehensive indexes obtained from the experiment, Acc, Pre and F1 are shown in the TABLE II.

TABLE II
Five-fold results of HCNNLDA using different classifiers

| Classifier | AUC | Acc | F1 | Pre | AUPR |
|---|---|---|---|---|---|
| SVM | 0.9589 | 0.8988 | 0.8993 | 0.8941 | 0.9543 |
| RF | 0.9549 | 0.8826 | 0.8825 | 0.8833 | 0.9532 |
| GBDT | 0.9437 | 0.8654 | 0.8623 | 0.8826 | 0.9424 |
| Adaboost | 0.9601 | 0.8880 | 0.8871 | 0.8802 | 0.9598 |
| **Xgboost** | **0.9752** | **0.9184** | **0.9194** | **0.9073** | **0.9740** |

According to the data in Figure 4 and TABLE II, it can be concluded that the performance of the XGB classifier is superior to other classifiers. The following conclusions can be drawn from the experimental results: (1) SVM is susceptible to data, and also depends on the selection of kernel function and parameter setting. (2) RF algorithm is greatly affected by noise, thus affecting the prediction results. (3) Adaboost, GBDT and XGB are ensemble learning methods. In the iteration process, the weights are updated according to the predicted results and used for the next round of learning. Different from the other two algorithms, XGB adds explicit regular terms, which helps to prevent overfitting and improve generalization ability. In conclusion, XGB achieved the best results compared to other classifiers.

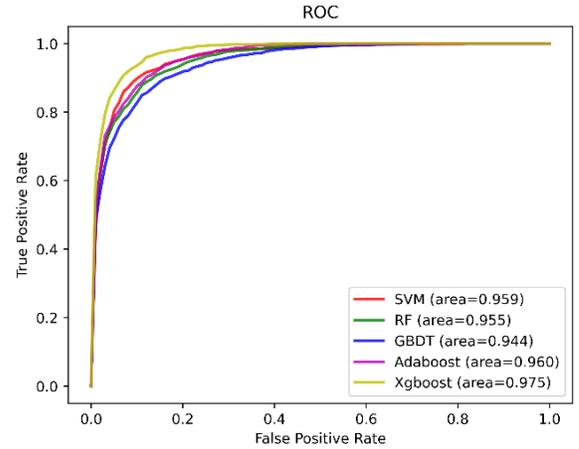

Figure 4. ROC curves of all classifiers.

### C. Comparative experiment

For the purpose of proving the predictive performance of HCNNLDA, it is compared with several methods: SIMCLDA [44], LDAP [45], Ping's method [27], LDASR [46], and GCNLDA[47]. The results are shown in Figure 5. With an AUC of 0.975, HCNNLDA shows the best experimental result. The factors that have achieved good experimental results are described as follows: Firstly, the potential features of lncRNA-disease association pairs are extracted by CNN. Second, the reference of the XGB classifier improves the predictive power. Therefore, it can be seen that HCNNLDA has more advantages than other methods in predicting LDAs.

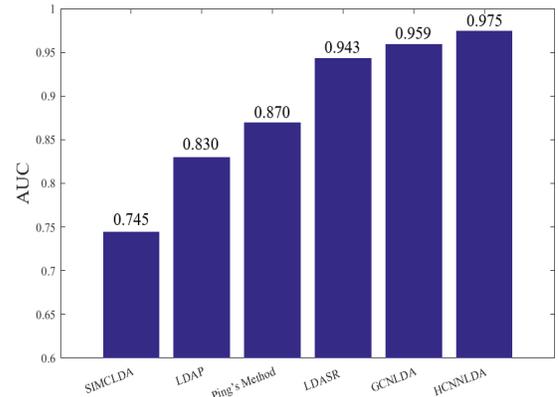

Figure 5. Comparison of AUC values for other methods.

*D. Case Study*

To evaluate the practicability and predictive capability of HCNNLDA, we selected cervical cancer, ovarian cancer, and prostate cancer as case studies. These cancers were chosen due to their significant impact and the availability of relevant data. The process involved sorting Long non-coding RNA-Disease Associations (LDAs) based on their associated scores and identifying the top 10 candidate LDAs for each cancer type.

To test the practicability and predictive ability of HCNNLDA, cervical cancer, ovarian cancer and prostate cancer are selected as disease case studies. According to their associated scores, sorting LDAs and collecting their respective 10 candidates. Since known LDAs are used to construct the model, Lnc2cancer [20] and MNDR [48] are used for validation. Here, I is used for Lnc2cancer, and II is used for MNDR.

The validation process involved cross-referencing the predicted candidate LDAs with entries in Lnc2Cancer and MNDR. This step was crucial to confirm the accuracy and relevance of our predictions. By leveraging these databases, we demonstrated that HCNNLDA could effectively predict potential LDAs, showcasing its practical application in identifying significant lncRNA-disease associations for cervical, ovarian, and prostate cancers. This approach not only underscores the model's predictive power but also its potential utility in guiding further biological research and clinical investigations.

The first selected disease is cervical cancer. As a common malignant tumor in gynecology, cervical cancer seriously harms women's physical and mental health. The correlation scores obtained from the predictions are arranged in descending order, and the top 10 prediction results are chosen. The results are shown in TABLE III. For example, SNHG16 is negatively associated with the expression level of miR-216-5p in cervical cancer cells. Studies have shown that there may be miR-101-3p binding site in SNHG16 sequence, so SNHG16 is effective in the treatment of cervical cancer. Besides, research has shown that TUG1 can promote the proliferation of cervical cancer cells, so inhibition of TUG1 expression can induce apoptosis of cervical cancer cells [49]. BANCR and LSINCT5 are not found to be related to cervical cancer in the two databases. However, related research have found that BANCR can inhibit the metastasis of cervical cancer and play a significant role in the treatment of diseases [50]. Similarly, LSINCT5 is found increased expression in cervical cancer cells [51].

TABLE III
Predicted LncRNAs for Cervical Cancer.

| Rank | lncRNA | Evidence |
|---|---|---|
| 1 | GAS5 | I; II |
| 2 | TUSC8 | I; II |
| 3 | SNHG16 | I; II |
| 4 | TUG1 | I; II |
| 5 | BANCR | Unconfirmed |
| 6 | HAGLR | II |
| 7 | PVT1 | I; II |
| 8 | HNF1A-AS1 | I; II |
| 9 | LSINCT5 | Unconfirmed |
| 10 | DANCR | I; II |

The second selected disease is ovarian cancer. As a kind of female cancer, the incidence of ovarian cancer ranks first among female malignant tumors. But if found in time, the chance of cure is higher. Similarly, the prediction result of ranking 10 is chosen for verification, and the results are shown in TABLE IV. In ovarian tumor samples, has been proved to be a low expression and has been proved to have the function of tumor suppressor gene [52, 53]. Besides, CCAT1 is significantly down-regulated in ovarian cancer cells, which can facilitate the proliferation and migration of ovarian cancer tissue. CCAT1 promotes the proliferation of ovarian cancer cells by interacting with miR-1290 in ovarian cancer cells [54]. In addition, research has concluded that CCAT1 could be a potential target for ovarian cancer treatment in the future [55]. In the two databases, BCAR4 has not been confirmed to be related to ovarian cancer. Based on the literature research, BCAR4 has a significant research value in the treatment of ovarian cancer [56].

TABLE IV
Predicted LncRNAs for Ovarian Cancer.

| Rank | lncRNA | Evidence |
|---|---|---|
| 1 | H19 | I; II |
| 2 | GAS5 | I; II |
| 3 | MEG8 | I; II |
| 4 | CCAT1 | II |
| 5 | DLEU2 | I |
| 6 | LINC00472 | I; II |
| 7 | BCAR4 | Unconfirmed |
| 8 | ZFAS1 | I; II |
| 9 | TUSC7 | I; II |
| 10 | CCAT1 | I; II |

The last selected disease for this study is prostate cancer, a malignant tumor whose incidence varies with age demographics. Among men, the mortality rate of prostate cancer is second only to lung cancer. However, if diagnosed at an early stage, it has a good prognosis due to the effectiveness of available therapies. For this study, the top 10 relevant results were selected, as shown in Table V. In prostate cancer samples, HOTTIP is negatively associated with miR-216a-5p expression levels, which promotes the proliferation of cancer cells through miR-216a-5p [57]. The down-regulation of HOTTIP can inhibit cancer, which has important reference value for the diagnosis of prostate disease [58]. In addition, MEG3 participates in the proliferation and migration of prostate cancer cells by regulating the expression of miR-9-5p [59]. In the two databases, DGCR5 has not been found to be related to prostate cancer. However, studies have shown that DGCR5 can inhibit prostate cancer [60]. These findings highlight the complex regulatory roles of various lncRNAs in prostate cancer, underscoring the potential for targeted therapies that modulate specific lncRNA interactions. The identification and understanding of these molecular mechanisms are crucial for developing more effective diagnostic and therapeutic strategies for prostate cancer.

TABLE V
Predicted LncRNAs for Prostate Cancer.

| Rank | lncRNA | Evidence |
|---|---|---|
| 1 | HOTTIP | I; II |
| 2 | GAS5 | I; II |
| 3 | PVT1 | I; II |
| 4 | MEG3 | I; II |
| 5 | PCA3 | I; II |
| 6 | FOXP4-AS1 | I; II |
| 7 | DGCR5 | Unconfirmed |
| 8 | HNF1A-AS1 | II |
| 9 | DLEU1 | II, |
| 10 | PCAT5 | I; II |

## V. Conclusion

Studying Long non-coding RNA-Disease Associations (LDAs) can significantly aid experts and scholars in the prevention and treatment of complex and challenging diseases. Traditional calculation methods for predicting LDAs have numerous limitations, making it imperative to develop novel approaches that can drive progress in various fields of biological research. In our research, we introduce a novel computational framework designed to enhance the understanding of candidate LDAs by leveraging Convolutional Neural Networks (CNN) and the Extreme Gradient Boosting (XGB) classifier. The framework operates through the integration of multiple types of heterogeneous data, leading to the construction of a heterogeneous network that includes Molecular-Disease Associations (MDAs), Long non-coding RNA-Disease Associations (LDAs), and Molecular-Long non-coding RNA-Disease Associations (MLDs). Utilizing the convolutional operations of CNNs, we extract low-dimensional feature representations of both lncRNAs and diseases, which are then input into an XGB classifier for the prediction of LDAs.

The efficacy of our proposed method, termed HCNNLDA, is validated through 5-fold cross-validation (CV) and comparison tests. These evaluations demonstrate the superior prediction performance of HCNNLDA in identifying potential LDAs. Additionally, case studies confirm the method's capability to predict previously unknown correlations, further showcasing its utility. In conclusion, our novel computational framework not only improves the prediction of potential LDAs but also holds promise for identifying numerous unknown associations. Future work aims to incorporate more relevant information into the prediction process, thereby enhancing the generalization ability of the model and contributing to more effective disease prevention and treatment strategies.